\pdfoutput=1

\documentclass[11pt]{article}

\usepackage{acl}

\usepackage{times}
\usepackage{latexsym}
\usepackage{multirow} 
\usepackage{float} 
\usepackage{makecell} 
\usepackage{enumitem}
\usepackage{balance}

\usepackage[T1]{fontenc}

\usepackage[utf8]{inputenc}

\usepackage{microtype}

\usepackage{graphicx}
\usepackage{latexsym,amsmath}
\usepackage{amssymb}
\usepackage{subfigure}
\usepackage{xspace}
\usepackage{color}
\usepackage{booktabs}

\newcommand{\model}{UPT\xspace}
\newcommand{\fullmodel}{Unified Prompt Tuning\xspace}

\title{Towards Unified Prompt Tuning for Few-shot Text Classification}

\author{Jianing Wang$^1$\thanks{\ \ \ J. Wang and C. Wang contributed equally to this work.}, Chengyu Wang$^2$\footnotemark[1], Fuli Luo$^2$, Chuanqi Tan$^2$, Minghui Qiu$^2$, Fei Yang$^3$,\\{\bf  Qiuhui Shi$^4$, Songfang Huang$^2$, Ming Gao$^1$\thanks{\ \ \ Corresponding author.}}\\
  $^1$ School of Data Science and Engineering, East China Normal University \\
  $^2$ Alibaba Group $^3$ Zhejiang Lab $^4$ Ant Group\\
  \texttt{lygwjn@gmail.com,\{chengyu.wcy,lfl259702,chuanqi.tcq\}@alibaba-inc.com}\\
  \texttt{minghui.qmh@alibaba-inc.com,yangf@zhejianglab.com}\\
  \texttt{qiuhui.sqh@antgroup.com,songfang.hsf@alibaba-inc.com}\\
  \texttt{mgao@dase.ecnu.edu.cn}
  }

\begin{document}
\maketitle
\begin{abstract}
Prompt-based fine-tuning has boosted the performance of Pre-trained Language Models (PLMs) on few-shot text classification by employing task-specific prompts. 
Yet, PLMs are unfamiliar with prompt-style expressions during pre-training, which limits the few-shot learning performance on downstream tasks.
It would be desirable if the models can acquire some prompting knowledge before adaptation to specific NLP tasks.
We present the~\emph{\fullmodel} (\emph{\model}) framework, leading to better few-shot text classification for BERT-style models by explicitly capturing prompting semantics from non-target NLP datasets.
In~\emph{\model}, a novel paradigm~\emph{Prompt-Options-Verbalizer} is proposed for joint prompt learning across different NLP tasks, forcing PLMs to capture task-invariant prompting knowledge.
We further design a self-supervised task named~\emph{Knowledge-enhanced Selective Masked Language Modeling} to improve the PLM's generalization abilities for accurate adaptation to previously unseen tasks.
After multi-task learning across multiple tasks, the PLM can be better prompt-tuned towards any dissimilar target tasks in low-resourced settings.
Experiments over a variety of NLP tasks show that~\emph{\model} consistently outperforms state-of-the-arts for prompt-based fine-tuning.
\footnote{All datasets are publicly available. Source codes will be released in EasyNLP~\cite{DBLP:journals/corr/abs-2205-00258}. URL:~\url{https://github.com/alibaba/EasyNLP}}
\end{abstract}

\begin{figure*}
\centering
\vspace{-.25em}
\includegraphics[width=0.975\textwidth]{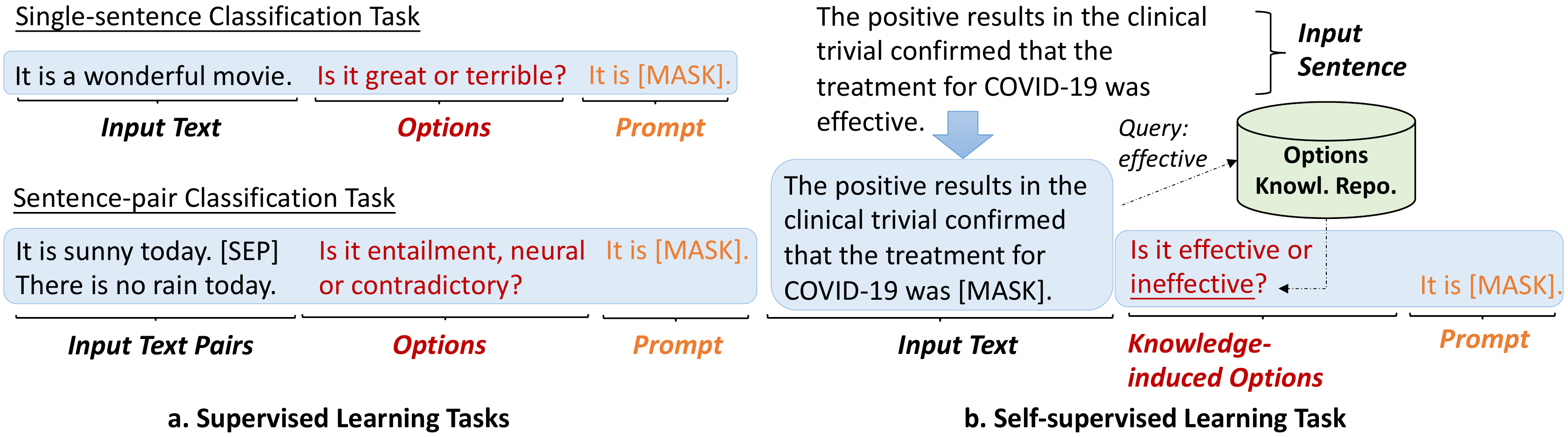}
\caption{\emph{\model} is a unified framework that learns prompting knowledge from non-target NLP datasets to improve the performance on target tasks, in the format of~\emph{Prompt-Options-Verbalizer} (Sect.~\ref{sec:framework}). Figures a) and b) show examples of supervised and self-supervised learning tasks (i.e.,~\emph{Knowledge-enhanced Selective MLM}, Sect.~\ref{sec:ss}).}
\label{fig:intro}
\vspace{-.5em}
\end{figure*}

\section{Introduction}

The emergence of Pre-trained Language Models (PLMs) has boosted the performance of a variety of NLP tasks~\cite{Qiu2020Pretrained, Xu2021Pretrained}. 
However, during fine-tuning, PLMs can perform poorly with few training samples due to model over-fitting~\cite{Gao2021Making}.

To alleviate this problem for low-resourced scenarios, {natural language prompts} have been applied to enable few-shot or zero-shot learning with PLMs~\cite{liu2021pre}. 
To make prompts more flexible and task-adaptive,~\emph{prompt tuning} freezes the PLM backbone and adjusts the representations of prompts~\cite{Lester2021the}. This type of method is especially suitable for ultra-large PLMs
that are difficult to tune. For BERT-style PLMs,~\emph{prompt-based fine-tuning} has been proposed, transforming text classification tasks into cloze-style problems~\cite{Schick2021Exploiting, Schick2021It, Gao2021Making}.
To specify, task-specific discrete templates with masked language tokens are added to input texts. The result tokens of the masked positions predicted by the Masked Language Modeling (MLM) head are used for class label prediction\footnote{For example, in the review analysis task, given an input ``It is a wonderful movie.'', one can add the prompt template ``Based on the review, it is [MASK].'' to the input. The output of the masked token ``great'' and ``terrible'' can be mapped to the positive and the negative class, respectively.}.
Therefore, the pre-trained knowledge acquired by PLMs can be better utilized by ``re-using'' the MLM training objective.
Witnessing the successful usage of prompts for few-shot learning, various following-up works have been conducted, such as {continuous prompt encoding}~\cite{Xiao2021GPT}, {knowledgeable prompt learning}~\cite{Hu2021Knowledgeable}, and {prompt generation}~\cite{Shin2020AutoPrompt}.

Recently, a few works~\cite{Wei2021Finetuned,Zhong2021Adapting, Mishra2021Reframing} focus on multi-task prompt tuning  on ultra-large PLMs. Specifically, they tune PLMs on full training samples from different tasks to force PLMs to learn more prompting knowledge, and directly make predictions over the target task by zero-shot learning.
Yet, we observe that for BERT-style PLMs, the performance is not satisfactory for two reasons.
1) These PLMs are sensitive to different designs of prompt templates and verbalizers~\cite{Xiao2021GPT}, which fail to adapt to target tasks with new prompts and verbalizers. 
2) There are word distribution differences between prompt-style texts and sentences in pre-training corpora.
It would be better if BERT-style PLMs can acquire some prompting knowledge before they are adapted to downstream tasks.
Therefore, a natural question arises:~\emph{how can we make BERT-style PLMs adapt to target NLP tasks accurately with more prompting knowledge?}

To address these issues, we introduce a novel framework named~\emph{\fullmodel} (\emph{\model}), facilitating better few-shot text classification performance for BERT-style models by explicitly capturing general prompting semantics from non-target datasets.
Specially, we propose a unified paradigm named~\emph{Prompt-Options-Verbalizer} (\emph{POV}), which enables mixture prompt-tuning over a series of~\emph{non-target NLP tasks} of varied types.
To further improve the model's generalization abilities on previously unseen tasks, we propose a novel auxiliary task named~\emph{Knowledge-enhanced Selective MLM} (\emph{KSMLM}), which mimics the behavior of MLM with explicit usage of prompts following the \emph{POV} paradigm.
After multi-task training is completed, the underlying PLM can be fine-tuned to fit any few-shot tasks using the same prompting paradigm.


In the experiments, we verify the effectiveness of~\emph{\model} over public NLP datasets of various tasks. Experimental results show that~\emph{\model} consistently outperforms state-of-the-art approaches for prompt-based few-shot fine-tuning. 
In summary, we make the following major contributions:
\begin{itemize}

\setlength{\itemsep}{0pt}

\item We introduce the novel~\emph{\model} framework to improve prompt-based fine-tuning for BERT-style models, which captures unified prompting semantics from multiple source tasks of various types for few-shot text classification on new target tasks.
    
\item In~\emph{\model}, a new paradigm~\emph{POV} is proposed for joint prompt tuning across different NLP tasks. We further design the self-supervised~\emph{KSMLM} task to improve the PLM's generalization abilities for accurate task adaptation.

\item Extensive experiments over various NLP datasets show that~\emph{\model} consistently outperforms state-of-the-arts for prompt-based few-shot fine-tuning by a relatively large margin. 
\end{itemize}

\section{\emph{\model}: The Proposed Framework}

We start with a brief overview of the~\emph{\model} framework, followed by its detailed techniques.

\subsection{A Brief Overview of~\emph{\model}}

For clarity, we introduce some basic notations. Let $\mathcal{D}^{*}$ be the $N$-way-$K$-shot training set of a target NLP task $\mathcal{T}^{*}$. The underlying PLM is parameterized by $\Theta$. The basic goal of few-shot learning is to obtain a high-performing model for $\mathcal{T}^{*}$ based on $\mathcal{D}^{*}$, with parameters initialized from $\Theta$.
As the size of $\mathcal{D}^{*}$ is only $N\times K$,
the model performance would be highly limited. Here, we assume that
there are $M$ other NLP tasks that are~\emph{dissimilar} to $\mathcal{T}^{*}$, i.e., $\mathcal{T}^{(1)},\cdots,\mathcal{T}^{(M)}$, with their (usually non few-shot) training sets denoted as $\mathcal{D}^{(1)},\cdots,\mathcal{D}^{(M)}$, respectively\footnote{Note that we constrain that $\mathcal{T}^{(1)},\cdots,\mathcal{T}^{(M)}$ are \emph{dissimilar} to $\mathcal{T}^{*}$ to deal with true low-resourced scenarios where no training sets of similar tasks are available.
If $\mathcal{T}^{(1)},\cdots,\mathcal{T}^{(M)}$ are similar to $\mathcal{T}^{*}$, one can directly apply transfer learning techniques to train the model, which is considered a relatively trivial problem and not the major focus of this work.
}.
The~\emph{\model} framework seeks to explore how to employ $\mathcal{D}^{(1)},\cdots,\mathcal{D}^{(M)}$ to enhance the performance of the PLM on a new task (such as $\mathcal{T}^{*}$) based on its own few-shot training set $\mathcal{D}^{*}$. 

In~\emph{\model}, the model is firstly trained over all the source tasks $\mathcal{T}^{(1)},\cdots,\mathcal{T}^{(M)}$, aiming to learn the semantics of prompts and the general methodology of solving downstream tasks by prompting. After that, it is prompt-tuned over a specific target task $\mathcal{T}^{*}$ in the low-resourced scenario.
To unify the learning process, each training sample $i$ in all different tasks (either $\mathcal{T}^{(1)},\cdots,\mathcal{T}^{(M)}$ or $\mathcal{T}^{*}$) is augmented in the same format, by means of the~\emph{Prompt-Options-Verbalizer} (\emph{POV}) triple $(P_i, O_i,V_i)$. Here, $P_i$ is the prompt. $O_i$ is the expression containing all possible options of the masked language token appearing in the prompt $P_i$ (i.e., the collection of label words). $V_i$ is the verbalizer that maps the target token predicted by the MLM head of the PLM to the class label.
Readers can also refer to the examples of supervised learning tasks in Figure~\ref{fig:intro}. 

In addition, we observe that the diversity of label words in original labeled tasks $\mathcal{T}^{(1)},\cdots,\mathcal{T}^{(M)}$ is limited. For previously unseen tasks, the optimization of these tasks alone often leads to a poorly generalized model that is biased towards these
tasks.
Therefore, we further introduce the self-supervised~\emph{Knowledge-enhanced Selective MLM} (\emph{KSMLM})
$\mathcal{\tilde T}$ as an auxiliary task. Specifically, take the sentences from source tasks training data $\mathcal{\tilde D}=\mathcal{D}^{(1)}\cup\mathcal{D}^{(2)}\cup\cdots\cup\mathcal{D}^{(M)}$ as inputs. These sentences are selectively masked, with options generated by rich knowledge mined from a massive corpus. An example is also in Figure~\ref{fig:intro}. 
Hence, the model has better generalization abilities and avoids catastrophic forgetting of pre-training knowledge.

\subsection{The Unified Prompting Paradigm}
\label{sec:framework}

A fundamental challenge for prompt-based training across $\mathcal{D}^{(1)},\cdots,\mathcal{D}^{(M)}$ for BERT-style models is that different NLP tasks have diverse sets of label words w.r.t. masked language tokens. When dealing with a mixture of training samples, a naive solution is to build a unified output prediction space, consisting of candidate label words from all tasks.
However, the enlarged output space makes it challenging for the PLM to optimize. 
Additionally, the output prediction space may not cover the label words of all possible unseen NLP tasks.

Here, we propose a unified prompting paradigm that augments each sample $i$ by  a~\emph{Prompt-Options-Verbalizer} (\emph{POV}) triple $(P_i, O_i,V_i)$.
$P_i$ is the prompt that provides task guidance (in line with PET~\cite{Schick2021Exploiting, Schick2021It}). $O_i$ is a fixed expression that explicitly provides selection for the model over all its candidate label words\footnote{Note that our framework is not restricted to binary classification. For NLP tasks with many labels, we can also directly list all the labels in options. More details and the external experiments can be found in Appendix.}. To facilitate the fast adaptation of arbitrary tasks, the verbalizer $V_i$ maps the output of the masked language token to the entire vocabulary $\mathcal{V}$. 
We can see that the options are 
crucial as they give strong indications on the possible outputs of the PLM (i.e., the candidates).
Overall, the output probability $q(v\vert i, P_i, O_i, \Theta)$ of the token $v\in\mathcal{V}$ w.r.t. the training sample $i$ is computed as follows:
\begin{equation*}
q(v\vert i, P_i, O_i, \Theta)=\frac{\exp({s(v\vert i, P_i, O_i, \Theta)})}{\sum_{v^{'}\in\mathcal{V}}\exp(s(v^{'}\vert i, P_i, O_i, \Theta))}
\vspace{-.4em}
\end{equation*}
where $s(v\vert i, P_i, O_i,\Theta)$ is the un-normalized score of the MLM head (before the softmax function) for generating token $v$ at the position of the masked language token with $i$,  $P_i$ and $O_i$ as inputs. Denote the entire prediction vector (of the length $\vert\mathcal{V}\vert$) as $Q(\mathcal{V}\vert i, P_i, O_i, \Theta)$. The~\emph{multi-task prompting loss} (denoted as $\mathcal{L}_{MP}$) can be written as follows:
\begin{equation*}
\begin{split}
\mathcal{L}_{MP}= & -\sum_{i\in\mathcal{D}}P(\mathcal{V}\vert i, P_i, O_i, \Theta)\cdot \\
& \log Q(\mathcal{V}\vert i, P_i, O_i, \Theta)
\end{split}
\vspace{-.4em}
\end{equation*}
where $\mathcal{D}=\bigcup_{k=1}^{M}{\mathcal{D}^{(k)}}$, and $P(\mathcal{V}\vert i, P_i, O_i, \Theta)$ is the one-hot ground-truth prediction vector.

In addition, we notice that  $\mathcal{D}^{(1)},\cdots,\mathcal{D}^{(M)}$ can be arbitrary labeled datasets with varied sizes. Optimizing $\mathcal{L}_{MP}$ directly on their original datasets would make the few-shot learner more likely to be biased towards larger datasets. In our work, we do stratified sampling to form a batch where a training sample $i$ from $\mathcal{D}^{(1)},\cdots,\mathcal{D}^{(M)}$ is picked with the probability proportional to its own dataset size (denoted as $w_i$), i.e.,
$
w_i=\frac{\log\vert\mathcal{D}^{(k)}\vert+\gamma}{M\cdot\gamma+\sum_{ k^{'}=1}^{M}\log\vert\mathcal{D}^{(k^{'})}\vert}
$
where $\gamma>0$ is a smoothing factor and $i\in \mathcal{D}^{(k)}$. Hence, we re-formulate $\mathcal{L}_{PT}$ as the~\emph{weighted multi-task prompting} (WMP) loss
$\mathcal{L}_{WMP}$:
\begin{equation*}
\begin{split}
\mathcal{L}_{WMP}= & -\sum_{i\in\mathcal{D}}w_i\cdot P(\mathcal{V}\vert i, P_i, O_i, \Theta)\cdot\\
& \log Q(\mathcal{V}\vert i, P_i, O_i, \Theta)
\end{split}
\vspace{-.4em}
\end{equation*}

\subsection{Extending Unified Prompting to Self-supervised Learning}
\label{sec:ss}

One drawback of the above approach is that the diversity of label words in these supervised learning tasks is usually limited, covering a narrow spectrum of the vocabulary $\mathcal{V}$. The model would not be well generalized for tasks with new label words. 
Hence, we leverage the idea of MLM pre-training, formulated by the~\emph{POV} paradigm.

\begin{figure}
\centering
\includegraphics[width=.9\linewidth]{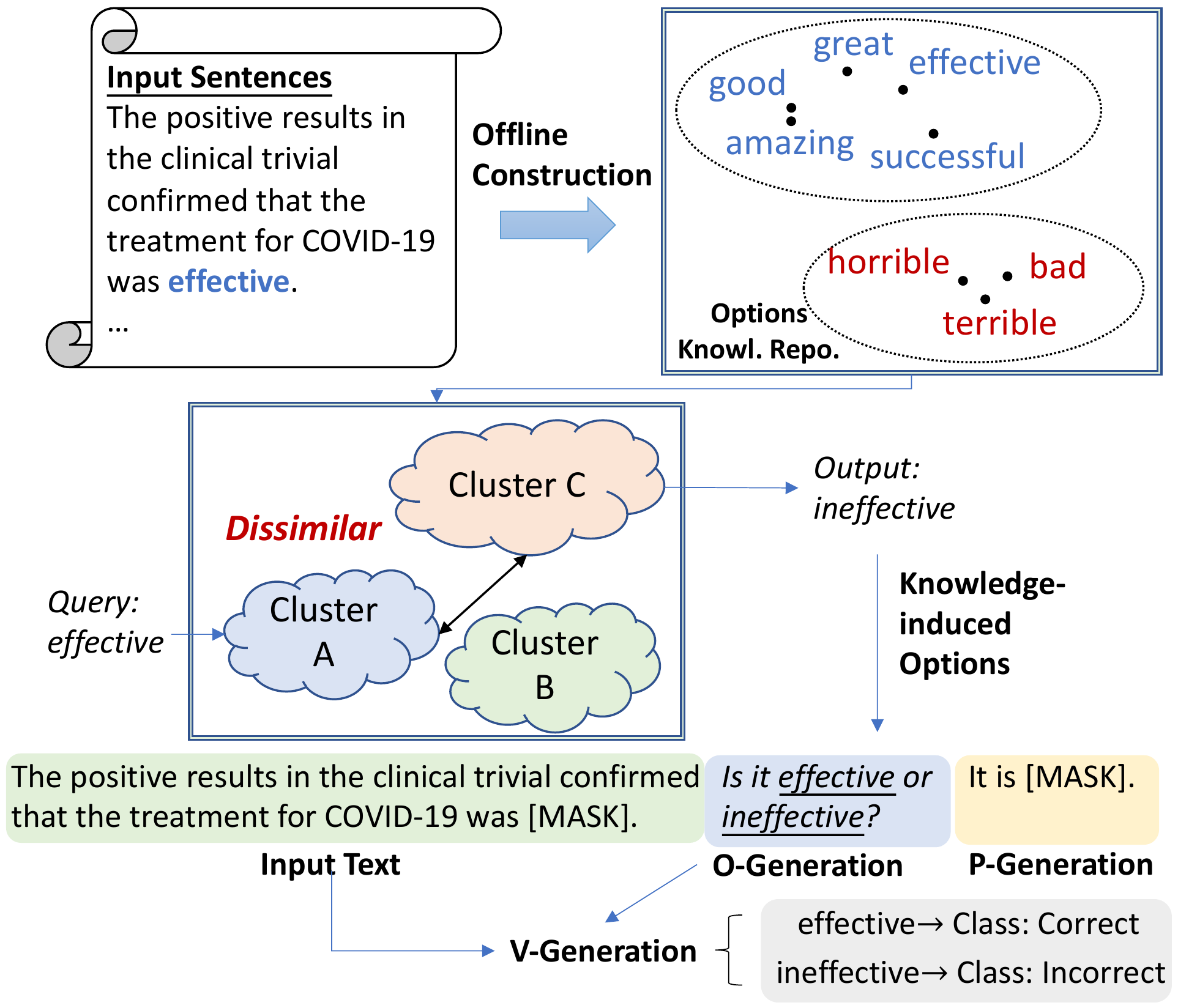}
\caption{An illustrated example of the~\emph{POV} generation process for the~\emph{KSMLM} task.
}
\vspace{-.5em}
\label{fig:verbalizer}
\end{figure}

As a naive approach, given a sentence, we can randomly mask a word and generate the options of the correct and a randomly selected word, and then ask the model to make the prediction.
Unfortunately, the seemingly feasible approach may ruin the training process, because not all words are suitable label words. For example, stop words and a large number of verbs and adverbs have not been used in any verbalizers in downstream tasks. The alternatives used in options should be reasonable, in order to make the model learn truly useful knowledge.
To address the issue, we present the self-supervised~\emph{KSMLM} task, with an example shown in Figure~\ref{fig:verbalizer}. In the following, we describe the~\emph{POV} construction process for~\emph{KSMLM}. After that, the loss function of the task is given.


\noindent\textbf{P-Generation.}
This process aims to generate a template with a \texttt{[MASK]} token for each sentence, which is fixed to be ``It is \texttt{[MASK]}.'' during the multi-task training stage. In the task-specific fine-tuning stage, we follow LM-BFF~\cite{Gao2021Making} to automatically generate templates for each task.
During training, the PLM is asked to predict the actual word of the masked position.

\noindent\textbf{O-Generation.}
From~\citet{Gao2021Making}, we can see that most label words for language understanding tasks are adjectives\footnote{In fact, we can also take into account the nouns if the label word space of target tasks is related to topics. Without loss of generality, we only consider adjectives in the experiments.} (such as ``great'' and ``terrible'' for sentiment analysis). Thus in our work, we detect all adjectives in the corpus by part-of-speech tagging models\footnote{We use the~\emph{spacy} toolkit in our work. URL:~\url{https://spacy.io/}.} and filter out low-frequency adjectives. The adjectives are then clustered by K-Means, with their token representations generated from the underlying PLM as features.
Formally, We construct a knowledge repository named~\emph{Options Knowledge Repository} (\emph{OKR}), in the form of triples $\mathcal{R}=\{(v,\vec{v},c_v)\}$, where $v$ is a candidate label word. $\vec{v}$ and $c_v$ denote the representation vector and the cluster membership of $v$, respectively. The cluster centroids are also stored.
We do not use existing lexicons such as WordNet~\cite{DBLP:journals/cacm/Miller95} because they may have limited coverage of label words. Additionally, the automatic process enables the extension of our algorithm to arbitrary languages and domains.

With the availability of~$\mathcal{R}$, we can generate knowledge-induced options.
Given a sentence with the masked word as $v$, we query $v$ against $\mathcal{R}$ for the most \emph{dissimilar} cluster w.r.t. $v$, denoted as $\tilde{c}_v$, where the cosine similarity of the vector representation $\vec{v}$ and the cluster centroid is employed as the similarity measure.
Finally, we randomly select one adjective from $\tilde{c}_v$ as the alternative label word to generate the~\emph{knowledge-induced options}.
The text expressions of options is fixed, i.e., ``Is it \texttt{[x1]} or \texttt{[x2]}?''. Readers can further refer to the example in Figure~\ref{fig:verbalizer}.

\noindent\textbf{V-Generation.}
For verbalizers, we map the true and the generated label words in the options to two classes, namely~\emph{Class: Correct} and~\emph{Class: Incorrect}. For instance, the verbalizers of the sample sentence in Figure~\ref{fig:verbalizer} are:
\begin{equation*}
\begin{split}
 \text{It is ``effective''.}\rightarrow & \text{``Class: Correct''} \\
 \text{It is ``ineffective''.}\rightarrow & \text{``Class: Incorrect''}
\end{split}
\end{equation*}


\noindent\textbf{Loss Function.}
The~\emph{KSMLM} loss is significantly different from the auxiliary MLM loss used in~\citet{Schick2021Exploiting, Schick2021It}. In~$\mathcal{\tilde{D}}$, each training sample $i$ can be directly extended to the training example for \emph{KSMLM} by \emph{POV} construction process with exactly one masked token, the~\emph{knowledge-induced options} $O_i$ and the prompt $P_i$. The PLM is trained to predict the correct masked word in the sentence, with the loss function:
$\mathcal{L}_{KSMLM}=-\sum_{i\in\mathcal{\tilde D}}P(\mathcal{V}\vert i, P_i, O_i, \Theta)\log Q(\mathcal{V}\vert i, P_i, O_i, \Theta)$. 
Overall, the loss function of~\emph{\model} $\mathcal{L}$ is defined as the summation of the WMP and KSMLM loss:
\begin{equation*}
\mathcal{L}=\mathcal{L}_{WMP}+\lambda\cdot\mathcal{L}_{KSMLM}
\vspace{-.4em}
\end{equation*}
where $\lambda\ge 0$ is the balancing hyper-parameter.

\noindent\textbf{Discussion.}
To our knowledge,
external knowledge has also been applied to other prompt-based methods, such as KPT~\cite{Hu2021Knowledgeable}. The major difference between KPT and ours is that~\emph{\model} uses the knowledge for options creation of the self-supervised task~\emph{KSMLM} that we proposed, in order to improve the model generalization abilities for accurate adaptation on new tasks. In contrast, previous works consider the expansion of verbalizers for specific downstream NLP tasks. 


\subsection{Few-shot Fine-tuning}
For a specific downstream task $\mathcal{T}^{*}$, the samples in the target few-shot training set $\mathcal{D}^{*}$ can be processed and computed in the same way as those supervised tasks used during~\emph{\model}. The learning consistency in the two stages ensures that the underlying PLM has already acquired prompting knowledge for $\mathcal{T}^{*}$. In addition, one can prompt-tune a single PLM over various tasks and uses it to fine-tune over any target tasks, making it computationally efficient to produce models for these applications. 

\section{Experiments}

\subsection{Experimental Settings}

In the experiments, we employ 9 public text classification datasets to evaluate the proposed~\emph{\model} framework, which are divided into three groups: sentiment analysis (Sentiment) (SST-2~\cite{Richard2013Recursive}, MR~\cite{Hu2004Mining}, CR~\cite{Pang2005Seeing}), Natural Language Inference (NLI) (MNLI~\cite{Williams2018A}, SNLI~\cite{Bowman2015A}, QNLI~\cite{wang2018GLUE}, RTE~\cite{dagan2005pascal}) and Paraphrase (MRPC~\cite{Dolan2005Automatically}, QQP\footnote{\url{https://www.quora.com/q/quoradata/}.}). The data statistics are shown in the Appendix. In default, $K=16$ (training instances per class).

\begin{table*}
\vspace{-.25em}
\centering
\begin{small}
\resizebox{\textwidth}{!}{
\begin{tabular}[\textwidth]{c | c | ccc | cccc | cc | c}
\midrule
\multirow{2}{*}{\textbf{Paradigm}} & \multirow{2}{*}{\textbf{Method}} & \multicolumn{3}{c |}{\textbf{Group 1: Sentiment.}} & \multicolumn{4}{c |}{\textbf{Group 2: NLI.}} & \multicolumn{2}{c |}{\textbf{Group 3: Paraphrase.}} & \multirow{2}{*}{\textbf{Avg.}} \\
 &  & \textbf{SST-2} & \textbf{MR} & \textbf{CR} & \textbf{MNLI} & \textbf{SNLI} & \textbf{QNLI} & \textbf{RTE} & \textbf{MRPC} & \textbf{QQP} & \\ 
\midrule
\multicolumn{12}{l}{\em Single-task methods w/o. the usage of dissimilar datasets ($K=16$)}\\
\midrule
FT & {Fine-tuning} & 81.1 \scriptsize$\pm$4.1 & 78.2\scriptsize $\pm$5.4 & 75.4\scriptsize $\pm$3.3 & 45.8\scriptsize $\pm$6.0 & 48.4\scriptsize $\pm$4.8 & 60.9\scriptsize $\pm$5.8 & 54.0\scriptsize $\pm$6.1 & 74.4\scriptsize $\pm$2.5 & 61.0\scriptsize $\pm$4.1 & 64.4\scriptsize $\pm$4.7 \\
\midrule
\multirow{5}{*}{PT} & PET & 91.8\scriptsize $\pm$1.3 & 86.4\scriptsize $\pm$2.9 & 90.5\scriptsize $\pm$1.9 & 58.4\scriptsize $\pm$2.2 & 59.4\scriptsize $\pm$2.9 & 61.3\scriptsize $\pm$1.8 & 65.7\scriptsize $\pm$2.0 & 74.5\scriptsize $\pm$1.6 & 67.6\scriptsize $\pm$3.1 & 72.8\scriptsize $\pm$2.2 \\
 & LM-BFF & 92.0\scriptsize $\pm$1.7 & 87.4\scriptsize $\pm$0.7 & 90.8\scriptsize $\pm$1.0 & 65.2\scriptsize $\pm$2.6 & \textbf{71.7}\scriptsize $\pm$4.9 & 69.1\scriptsize $\pm$2.8 & 69.5\scriptsize $\pm$2.0 & 74.2\scriptsize $\pm$2.3 & 63.5\scriptsize $\pm$1.2 & 75.9\scriptsize $\pm$2.4 \\
 & P-Tuning & 92.6\scriptsize $\pm$1.6 & 87.0\scriptsize $\pm$1.2 & 91.7\scriptsize $\pm$1.4 & 62.4\scriptsize $\pm$2.3 & 70.2\scriptsize $\pm$2.1 & 68.8\scriptsize $\pm$3.5 & \textbf{70.8}\scriptsize $\pm$2.5 & 73.4\scriptsize $\pm$1.9 & 67.6\scriptsize $\pm$0.8 & 76.0\scriptsize $\pm$1.6 \\
 & PPT & 92.3\scriptsize $\pm$0.5 & 87.1\scriptsize $\pm$1.6 & 90.9\scriptsize $\pm$1.3 & 64.9\scriptsize $\pm$2.0 & 71.4\scriptsize $\pm$1.5 & 68.8\scriptsize $\pm$2.9 & 67.9\scriptsize $\pm$2.6 & 74.8\scriptsize $\pm$2.1 & 67.2\scriptsize $\pm$1.2 & 76.1\scriptsize $\pm$1.8\\
 & \bf\emph{\model}-Single & \textbf{92.9}\scriptsize $\pm$1.0 & \textbf{87.7}\scriptsize $\pm$1.5 & \textbf{91.8}\scriptsize $\pm$0.7 & \textbf{65.6}\scriptsize $\pm$1.4 & 71.2\scriptsize $\pm$2.3 & \textbf{70.1}\scriptsize $\pm$1.6 & 68.9\scriptsize $\pm$1.7 & \textbf{75.1}\scriptsize $\pm$0.9 & \textbf{72.1}\scriptsize $\pm$2.0 & \textbf{77.2}\scriptsize $\pm$1.5 \\
\midrule
\multicolumn{12}{l}{\em Multi-task methods w. the usage of dissimilar datasets ($K=16$)}\\
\midrule
\multirow{5}{*}{PT} & MT\scriptsize (Zero-shot) & 58.7\scriptsize $\pm$1.6 & 59.0\scriptsize $\pm$3.6 & 58.9\scriptsize $\pm$2.8 & 36.3\scriptsize $\pm$3.3 & 39.2\scriptsize $\pm$3.2 & 40.9\scriptsize $\pm$2.5 & 54.9\scriptsize $\pm$1.4 & 70.6\scriptsize $\pm$2.6 & 42.8\scriptsize $\pm$2.5 & 51.3\scriptsize $\pm$2.2 \\
 & MT\scriptsize (Few-shot) & 92.1\scriptsize $\pm$1.4 & 86.5\scriptsize $\pm$1.3 & 91.0\scriptsize $\pm$2.2 & 69.6\scriptsize $\pm$1.1 & 67.1\scriptsize $\pm$2.7 & 68.9\scriptsize $\pm$2.3 & 68.6\scriptsize $\pm$1.2 & 71.0\scriptsize $\pm$1.4 & 74.8\scriptsize $\pm$2.1 & 76.7\scriptsize $\pm$1.7 \\
 & \emph{\model}\scriptsize (Zero-shot) & 74.5\scriptsize $\pm$1.2 & 73.9\scriptsize $\pm$1.3 & 72.4\scriptsize $\pm$1.4 & 43.7\scriptsize $\pm$2.0 & 46.0\scriptsize $\pm$2.1 & 53.9\scriptsize $\pm$1.9 & 57.1\scriptsize $\pm$1.0 & 70.7\scriptsize $\pm$0.9 & 56.5\scriptsize $\pm$1.3 & 61.0\scriptsize $\pm$1.5 \\
 & \bf\emph{\model} & \textbf{93.5}\scriptsize $\pm$0.6 & 88.1\scriptsize $\pm$0.9 & 91.4\scriptsize $\pm$1.2 & 70.1\scriptsize $\pm$1.4 & 68.2\scriptsize $\pm$1.2 & 69.9\scriptsize $\pm$1.5 & 73.5\scriptsize $\pm$1.5 & \textbf{77.0}\scriptsize $\pm$1.1 & 78.8\scriptsize $\pm$1.7 & 78.9\scriptsize $\pm$1.4 \\
 & \bf\emph{\model}-SE & 93.1\scriptsize $\pm$0.4 & \textbf{88.4}\scriptsize $\pm$0.9 & \textbf{92.1}\scriptsize $\pm$1.0 & \textbf{71.4}\scriptsize $\pm$1.1 & \textbf{73.6}\scriptsize $\pm$0.6 & \textbf{70.5}\scriptsize $\pm$1.6 & \textbf{75.8}\scriptsize $\pm$0.8 & 76.2\scriptsize $\pm$0.4 & \textbf{79.6}\scriptsize $\pm$1.3 & \textbf{80.1}\scriptsize $\pm$1.1 \\
\midrule
\end{tabular}
}
\end{small}
\vspace{-.25em}
\caption{\label{tab:general_results}Comparison between~\emph{\model} and baselines over all testing sets in terms of accuracy (\%) and standard deviation. ``FT'' and ``PT'' refer to the \emph{fine-tuning} and \emph{prompt-based fine-tuning} paradigm, respectively.
The methods in bold refer to our approach and its variants.
The scores of baselines are re-produced using their open-source codes.}
\vspace{-.5em}
\end{table*}

As mentioned above, during~\emph{\model}, we only leverage full training data from all \emph{dissimilar} task groups, and then prompt-tune the model on the target task in the low-resourced setting. 
For example, when the target task is SST-2, the training data during~\emph{\model} is from NLI and Paraphrase. 
The underlying PLM is the RoBERTa-large model (with 335M parameters)~\cite{Liu19RoBERTa}, unless otherwise specified.
The baselines include standard \emph{fine-tuning}, and four recently proposed few-shot learning algorithms: PET~\cite{Schick2021Exploiting}\footnote{\url{https://github.com/timoschick/pet}}, LM-BFF~\cite{Gao2021Making}\footnote{For a fair comparison with other approaches, we train the underlying models by LM-BFF with manual-compiled prompts without demonstration learning. URL: \url{https://github.com/princeton-nlp/LM-BFF}}, P-tuning~\cite{Xiao2021GPT}\footnote{\url{https://github.com/THUDM/P-tuning}} and PPT~\cite{Gu2021PPT}. To make a fair comparison with these single-task baselines, a variant of our approach (denoted as~\emph{\model}-Single) is also implemented by only fine-tuning over the few-shot target task based on~\emph{POV} without the usage of \emph{dissimilar} supervised source datasets.

As we use other \emph{dissimilar} datasets to train our model, we also include two multi-task methods that are~\emph{meta-tuned} using the same \emph{dissimilar} datasets as strong baselines, namely MT (Zero-shot) and MT (Few-shot)~\cite{Zhong2021Adapting}\footnote{In~\citet{Zhong2021Adapting}, the authors only conduct zero-shot learning using larger PLMs. To make their work comparable to ours,  we re-implement their algorithm over the Roberta model on our datasets under two settings. MT (Zero-shot) refers to the model tuned only using \emph{dissimilar} full datasets. MT (Few-shot) further tunes the entire model over the target few-shot training set based on the prompts.
Note that a few contemporaneous works (such as~\citet{Wei2021Finetuned}) also consider multi-task zero-shot learning. Because the settings and model scales are significantly different from ours, they are not directly comparable.
}. We also implement the zero-shot version of~\emph{\model}, denote as \emph{\model} (Zero-shot). 
In addition, given a supervised NLP task, multiple prompts can be manually crafted. 
By augmenting one training sample with these prompts, we can automatically realize~\emph{self-ensemble learning}. 
For the self-ensemble version of~\emph{\model}, we employ five different prompts. For each input sample, we randomly select one expression of options and one set of verbalizers. We denote this method as~\emph{\model}-SE. 
The designed prompts, options, and verbalizers are listed in the Appendix. All the results of these models are evaluated in terms of averaged accuracy and its standard deviation, over 5 random seeds.


Our~\emph{\model} framework is implemented in PyTorch and run with NVIDIA V100 GPUs. Specifically, we train our model with the Adam optimizer. The learning rate for all training stages is fixed to be 1e-5. We set the default hyper-parameters as $\gamma=0.001$ and $\lambda=0.1$, which are also tuned over the development sets. The parameter regularizers are the same as in~\citet{Gao2021Making}.


\begin{table}
\centering
\begin{small}
\resizebox{\linewidth}{!}{
\begin{tabular}{ c | ccc | c}
\midrule
\textbf{BERT Scale} & \textbf{SST-2} & \textbf{MR} & \textbf{CR} & \textbf{Avg.} \\
\midrule
Base & 82.6\scriptsize $+$3.8 & 71.1\scriptsize $+$9.3 & 78.1\scriptsize $+$8.9 & 77.2\scriptsize $+$7.3 \\
Medium & 68.0\scriptsize $+$3.0 & 63.4\scriptsize $+$4.2 & 70.2\scriptsize $+$6.1 & 67.2\scriptsize $+$4.4 \\
Small & 66.3\scriptsize $+$3.7 & 58.1\scriptsize $+$4.6 & 68.2\scriptsize $+$5.5 & 64.2\scriptsize $+$4.6 \\
Mini & 58.8\scriptsize $+$3.1 & 59.4\scriptsize $+$7.6 & 65.8\scriptsize $+$7.5 & 61.3\scriptsize $+$6.1 \\
Tiny & 54.2\scriptsize $+$3.8 & 54.0\scriptsize $+$1.3 & 54.4\scriptsize $+$5.2 & 54.2\scriptsize $+$3.4 \\
\midrule
\end{tabular}
}
\end{small}
\caption{\label{tab:scale}Results of model scale analysis. We report the accuracy (\%) of \emph{\model} based on BERT with other scales, and relative improvements, compared to the models w/o. prompt learning over \emph{dissimilar} datasets.
}
\vspace{-.25em}
\end{table}

\subsection{Main Results}

In Table~\ref{tab:general_results}, we report the general experimental results of~\emph{\model} and all the baselines.
The results show that: 1) Prompt-based methods (i.e., PET~\cite{Schick2021Exploiting}, LM-BFF~\cite{Gao2021Making}, P-tuning~\cite{Xiao2021GPT} and PPT~\cite{Gu2021PPT}) have large improvements over standard \emph{fine-tuning}.
2)~\emph{\model}-Single outperforms previous few-shot learning models in average, which indicates that the utilization of \emph{POV} is better than vanilla prompts~\cite{Schick2021Exploiting}.
3) \emph{\model} (both the vanilla and the ensemble version) consistently outperforms all baselines on all tasks,  which demonstrates that our framework possesses better generalization by learning from \emph{dissimilar} groups of tasks\footnote{We also conduct the single-tail paired t-test to compare our approach against few-shot baselines across tasks. The result is $p<0.05$, indicating the statistical significance.}.
4) MT (Zero-shot)~\cite{Zhong2021Adapting} and \emph{\model} (Zero-shot) do not yields satisfactory results on BERT-style models. Different from ultra-large models, we suggest that few-shot prompt-tuning is necessary for BERT-style models to produce good results over these tasks. 
5) By comparing~\emph{\model} against MT (Few-shot), we can see that the proposed~\emph{POV} paradigm and the self-supervised~\emph{KSMLM} task are more effective for few-shot learning.
6) Generally, \emph{\model}-SE improves the averaged accuracy on all tasks by 1.2\% than \emph{\model}. It means that self-ensemble learning can enhance model generalization, but the improvement is not consistent across all tasks.
A possible cause is that some prompts and options are not optimal for the target task.

\begin{figure}[ht]
\centering
\begin{tabular}{cc}
\begin{minipage}[t]{0.45\linewidth}
    \includegraphics[width = 1\linewidth]{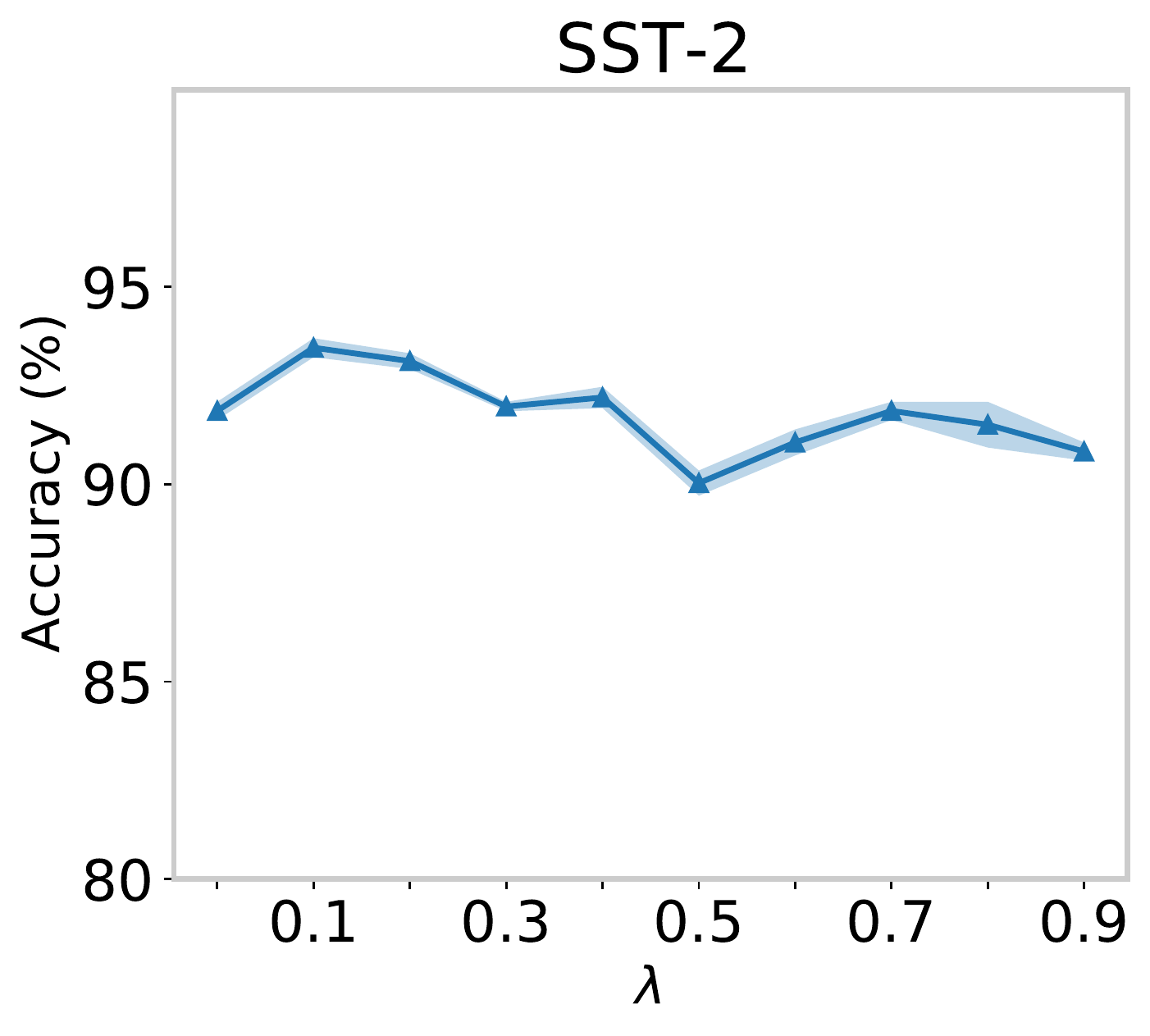}
\end{minipage}
\begin{minipage}[t]{0.45\linewidth}
    \includegraphics[width = 1\linewidth]{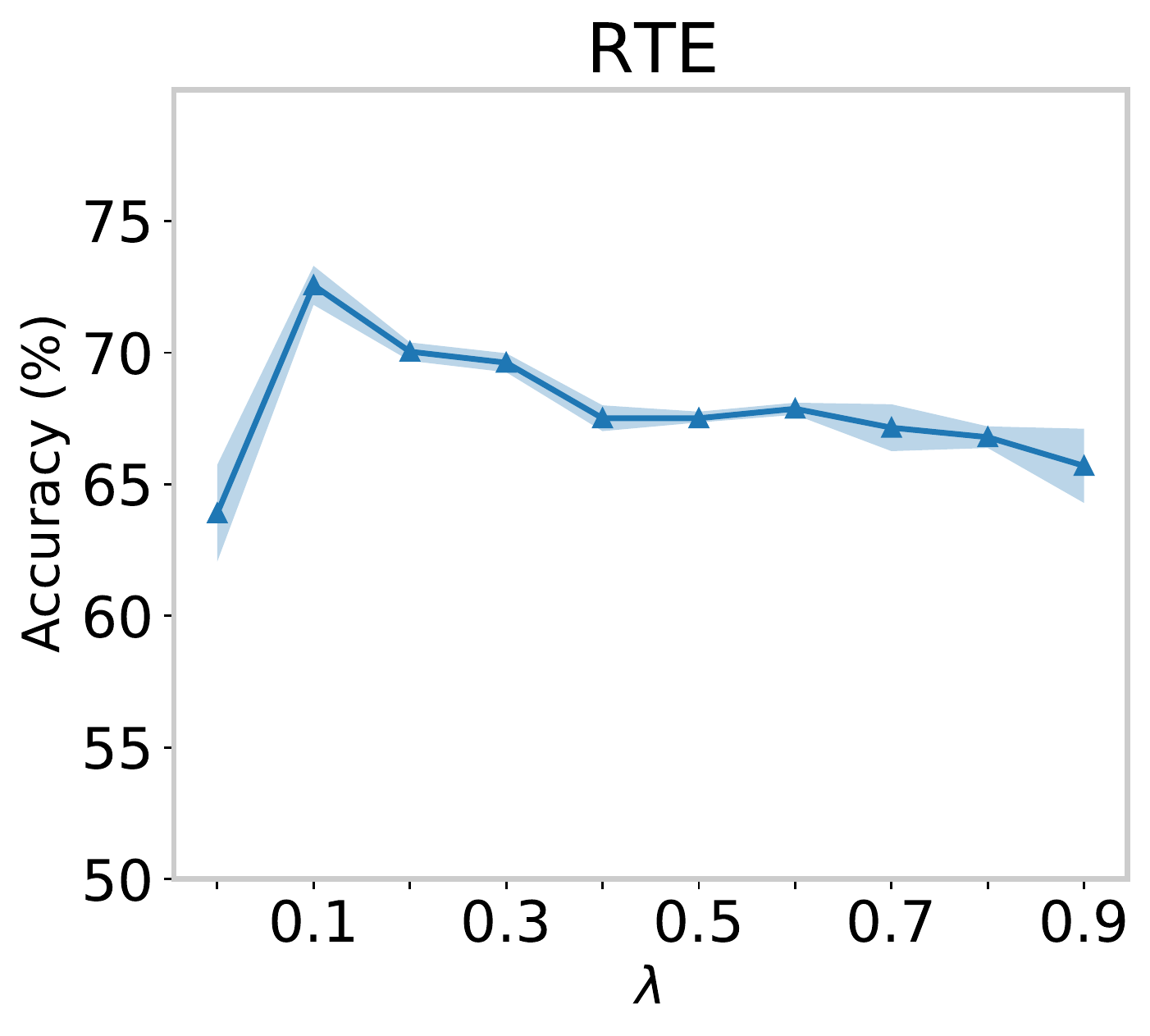}
\end{minipage}
\end{tabular}
\vspace{-.25em}
\caption{Parameter analysis w.r.t. hyper-parameter $\lambda$.}
\label{fig:para}
\vspace{-.25em}
\end{figure}

\subsection{Model Analysis}

\noindent\textbf{Parameter Analysis.}
We conduct parameter analysis to investigate the best choice of the balance coefficient $\lambda$. Results over SST-2 and RTE are shown in Figure~\ref{fig:para}. We have the best performance when $\lambda=0.1$, which indicates that our proposed~\emph{\model} possess generalization when it is jointly trained over the self-supervised~\emph{KSMLM} task. We also observe that the performance decreases when $\lambda$ becomes larger. 
This means \emph{KSMLM} is a suitable regularization task, but also may introduce a lot of prompts and options that are irrelevant to downstream tasks. This opens up new opportunities for model improvement.


\begin{table}
\centering
\begin{small}
\resizebox{\linewidth}{!}{
\begin{tabular}{l | ccc}
\midrule
\textbf{Method}/\textbf{Group} & \textbf{Group 1} & \textbf{Group 2} & \textbf{Group 3}\\
\midrule
MT \scriptsize (Few-shot) & 89.9 & 68.6 & 72.9 \\
~\emph{\model} & \textbf{91.0} & \textbf{70.2} & \textbf{77.9}\\
\midrule
\quad w/o. ~\emph{POV} & 90.2 & 68.9 & 74.2 \\
\quad w/o. ~\emph{KSMLM} & 90.9 & 69.1 & 73.7 \\
\quad w/o. ~\emph{POV}\&\emph{KSMLM} & 89.6 & 68.7 & 73.5 \\
\quad w/o. ~\emph{OKR} & 90.7 & 69.9 & 76.8 \\
\midrule
\end{tabular}
}
\end{small}
\vspace{-.5em}
\caption{\label{tab:ablation}Ablation study in terms of accuracy (\%). Standard deviations are omitted here to save space.
}
\vspace{-.25em}
\end{table}

\noindent\textbf{Ablation Study.}
To clearly verify the contributions of each component in~\emph{\model}, we conduct an ablation study over all groups and report the mean accuracy. As shown in Table~\ref{tab:ablation}, w/o. \emph{POV} denotes the method with manually designed prompts without the usage of any options.
w/o. \emph{KSMLM} equals the setting with $\lambda=0$, which is the same as \emph{\model}-Single.
w/o. \emph{OKR} means that we randomly choose the alternative label words in the options without knowledge guidance when we optimize the~\emph{KSMLM} task.
w/o. \emph{POV} \& \emph{KSMLM} denotes the method without any options and the auxiliary \emph{KSMLM} task.
The results show that no matter which module is removed, the model performance is affected.
Particularly, when we remove both \emph{POV} and \emph{KSMLM}, the performance is decreased by 1.4\%, 1.5\%, 4.4\%, respectively. The accuracy values of this setting are lower than w/o. \emph{POV} and w/o. \emph{KSMLM}, which suggests that both two components highly contribute to the high performance of our framework. 
We also find that w/o. ~\emph{POV} or w/o. \emph{KSMLM} both outperform MT (Few-shot) over all groups.  Additionally, we find that if we use \emph{KSMLM} but remove \emph{OKR}, the results decrease over all these tasks, but are still higher than w/o. \emph{KSMLM}. It means that the options knowledge that we mine from the corpus is suitable for the self-supervised learning task.

\begin{figure}[t]
\centering
\begin{tabular}{cc}
\begin{minipage}[t]{0.45\linewidth}
    \includegraphics[width = 1\linewidth]{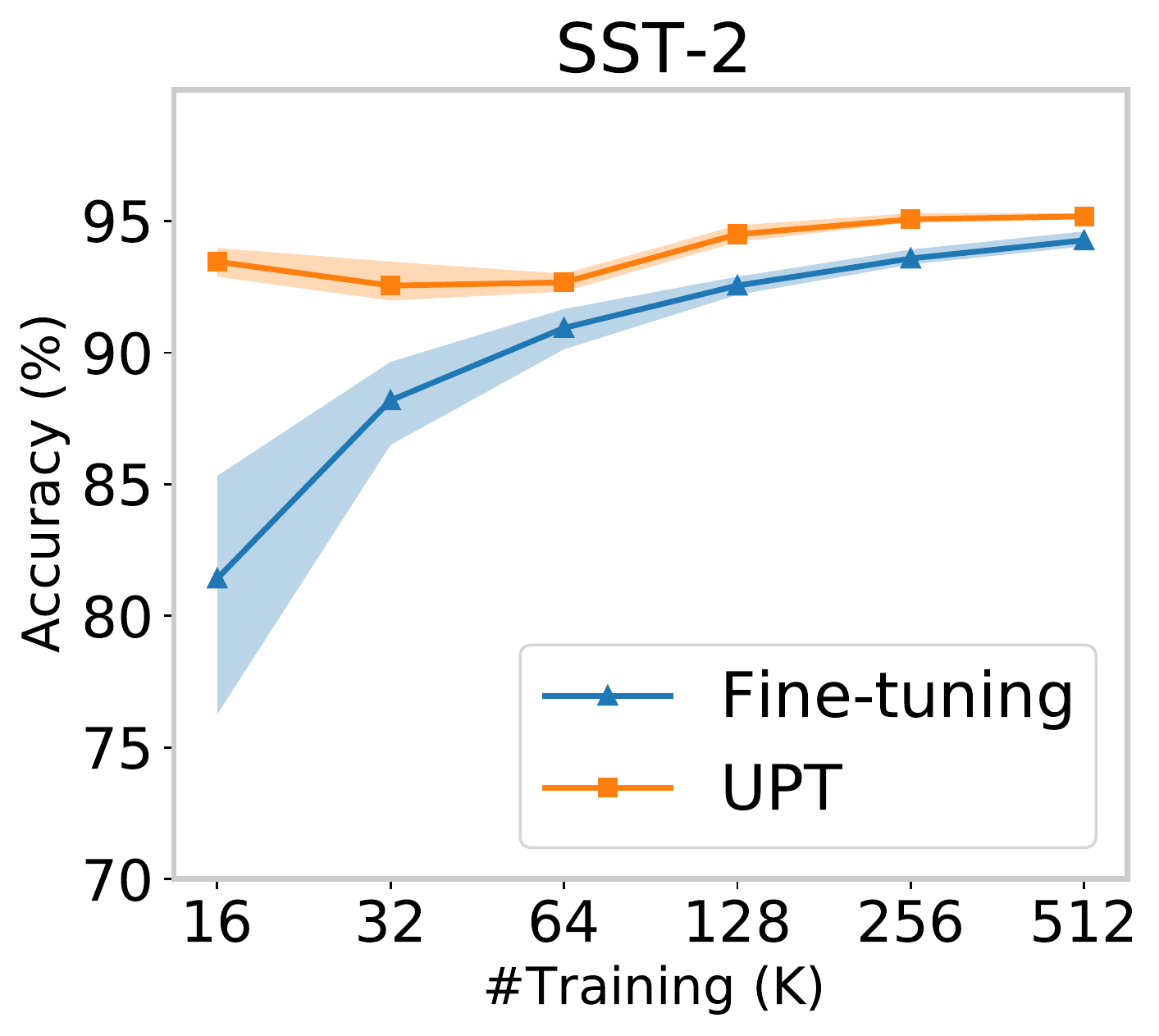}
\end{minipage}
\begin{minipage}[t]{0.45\linewidth}
    \includegraphics[width = 1\linewidth]{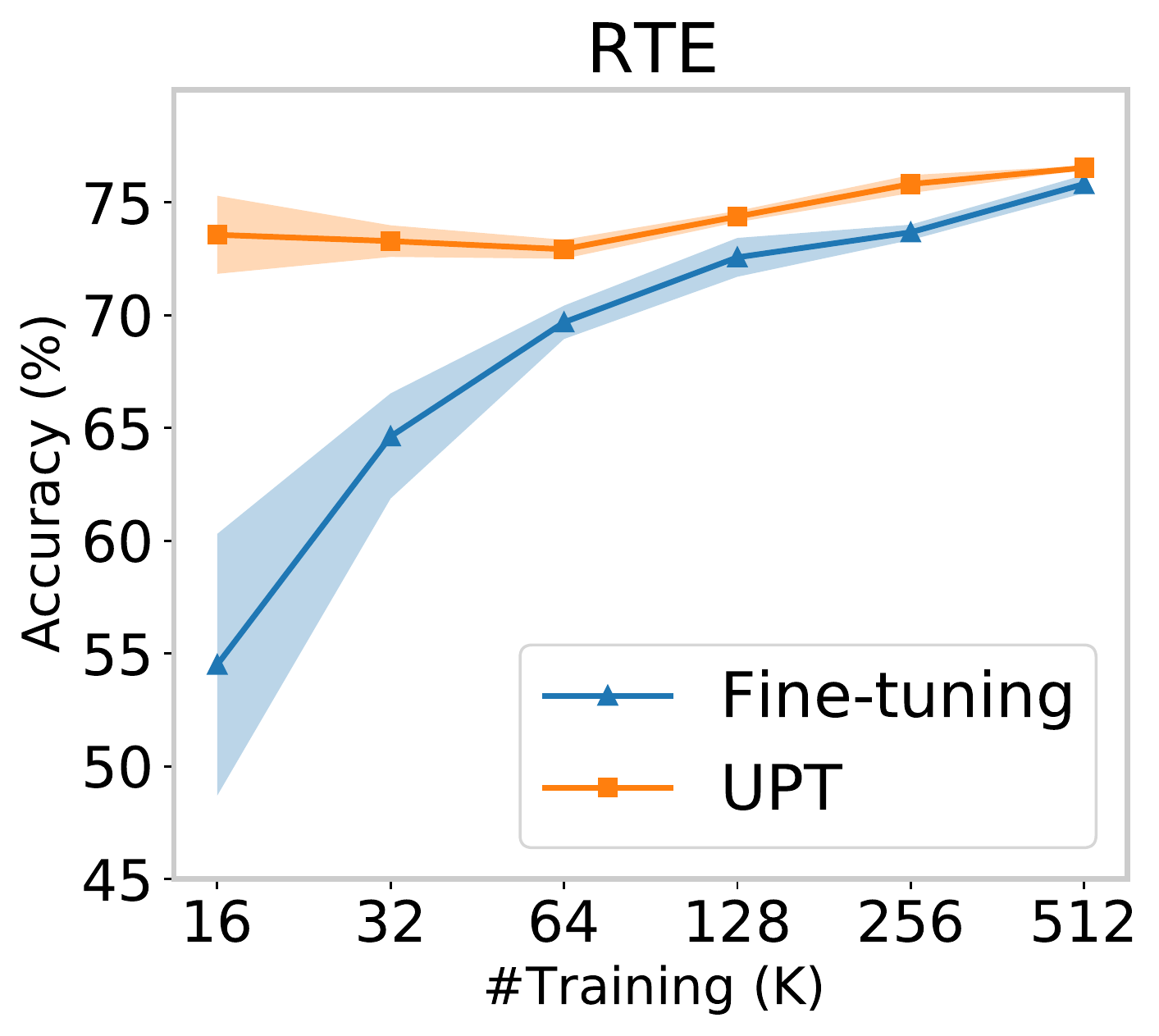}
\end{minipage}
\end{tabular}

\caption{Results of sample efficiency analysis. We compare~\emph{\model} with standard \emph{fine-tuning} with different numbers of training samples $K$ over two tasks.}
\label{fig:sample_efficiency}
\vspace{-.25em}
\end{figure}

\noindent\textbf{Sample Efficiency.}
We further explore the model effects with different numbers of training samples per class ($K$) from 16 to 512. We also use standard \emph{fine-tuning} as the reference. As shown in Figure~\ref{fig:sample_efficiency}, each point refers to the averaged score across 5 randomly sampled datasets. We observe that our~\emph{\model} consistently achieves higher scores regardless of the number of training samples. In addition, the variance of \emph{\model} is lower than \emph{fine-tuning}, meaning that the stability of our method is better. This is different from other prompt-based methods~\cite{Schick2021Exploiting, Schick2021It, Gao2021Making}.

\noindent\textbf{Model Scale Analysis.}
To further show that~\emph{\model} can improve the model performance regardless of the scales, we regard multiple small-scale BERT as model backbones\footnote{\url{https://github.com/google-research/bert}}.
Due to space limitations, we only illustrate the results in Table~\ref{tab:scale} over SST-2, MR, and CR.
To make a fair comparison, we also test the performance without the usage of \emph{dissimilar} NLP datasets and show the relative improvements.
The results demonstrate that the model scale plays an important role in the ability of model generalization.
We also find that \emph{\model} that uses \emph{dissimilar} datasets can highly improve the effectiveness, especially on small-scale PLMs.
Therefore, our method is better suitable for producing high-performing small PLMs for online applications.


\begin{figure}[t]
\centering
\includegraphics[width=\linewidth]{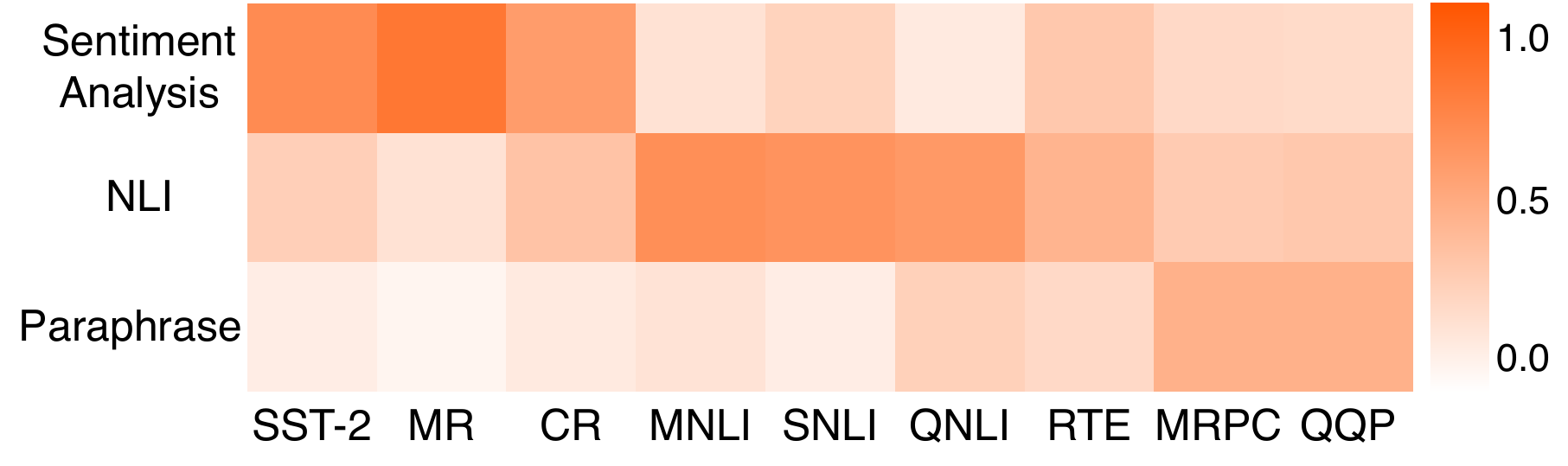}
\caption{Adaptation efficiency between task groups. The shade of color indicates the degree of adaptation.}
\label{fig:group_efficiency}
\end{figure}

\noindent\textbf{Adaptation Efficiency of Task Groups.}
Because we focus on multi-task training before prompt-tuning over the target task in low-resourced settings. Therefore, it is worth exploring which/how many groups of tasks have a better effect on the adaptation improvement. 
Specifically, when given a target task (e.g., MNLI), we only choose one group of tasks (e.g., MRPC and QQP of Group 3 (Paraphrase)) for 
multi-task prompt-tuning, and then fine-tune the model on the target task. As shown in Figure~\ref{fig:group_efficiency}, the cell in the $i$-th row and $j$-th column denotes the relative improvement from single-task learning over the $j$-th task to the setting where the $i$-th group is added for multi-task prompt learning. For visualization, we normalize the values of each column to show the percentage of influence of each group. The results show that the performance of a target task improves the most when we add data samples from other datasets within the same task group. However, in low-resourced scenarios, similar datasets are not available. By using~\emph{\model}, we can even transfer the knowledge from datasets from \emph{dissimilar} tasks to the target task.

\begin{figure}[t]
\centering
\begin{tabular}{c}
\begin{minipage}[t]{0.75\linewidth}
    \includegraphics[width = 1\linewidth]{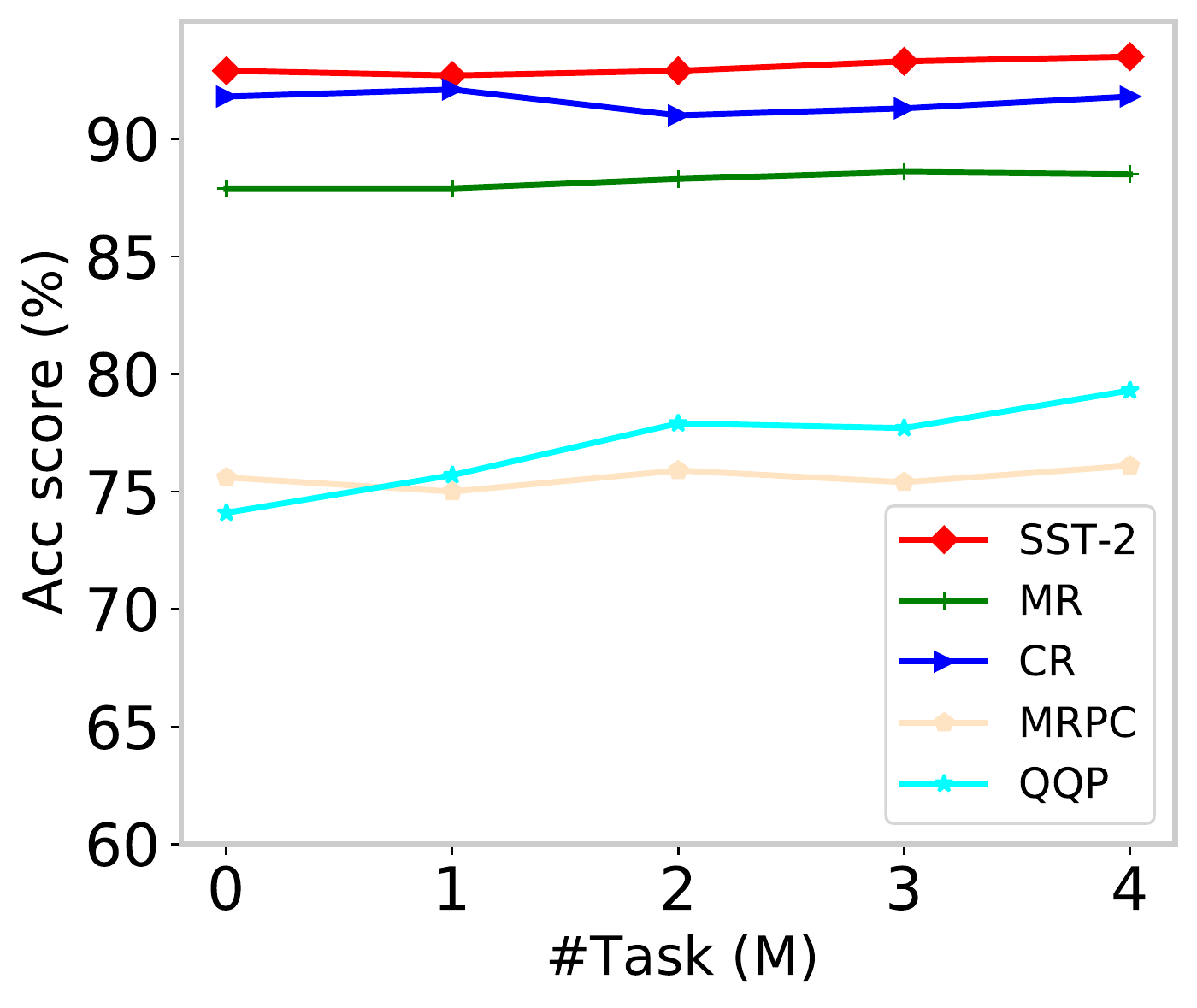}
\end{minipage}
\end{tabular}
\caption{Adaptation efficiency between the different numbers of NLI tasks ($M$) and each target task from Sentiment and Paraphrase.}
\label{fig:adaptation_efficiency}
\end{figure}

Specifically, taking NLI as the source group, we randomly choose $M$ dataset(s) from the group as our source tasks and then prompt-tune the model on each target task. The results from Figure~\ref{fig:adaptation_efficiency} demonstrate that the accuracy is further improved when we increase the value $M$. We also find that the improvements over MRPC and QQP are more obvious. We suggest that NLI is easier to be adapted to paraphrase tasks because they both model the relations between sentence pairs.

\section{Related Work}


\noindent\textbf{Pre-trained Language Models.}
Recently, benefited from the powerful modeling abilities of PLMs and computational resources, we have witnessed the qualitative improvement of multiple NLP tasks~\cite{Qiu2020Pretrained, Xu2021Pretrained}. For examples, the large GPT model series~\cite{radford2019language,Brown2020Language} utilizes multi-layer transformer decoders to capture left-to-right semantics of natural languages.  BERT~\cite{Devlin2019BERT} focuses on the learning of bidirectional contextual representations.
Other notable PLMs include Transformer-XL~\cite{Dai2019TransformerXL}, ELMo~\cite{Peters18Deep}, RoBERTa~\cite{Liu19RoBERTa}, AlBERT~\cite{Lan20ALBERT},
XLNet~\cite{Yang19XLNet}, StructBERT~\cite{wang2019structbert}, 
T5~\cite{Raffel2020Exploring}, etc. As model architecture is not the focus of our work, we do not elaborate. 

\noindent\textbf{Prompt-based Learning.}
Fine-tuning PLMs directly by learning the \texttt{[CLS]} head may perform poorly with few training samples~\cite{liu2021pre}. Recently, the huge GPT-3 model~\cite{Brown2020Language} has been proposed to enable in-context learning, which introduces handcrafted prompts and demonstrations.
\citet{Schick2021Exploiting} apply handcrafted prompts to {prompt-based fine-tuning} for BERT-style models.
To facilitate the automatic prompt generation, \citet{Gao2021Making} present LM-BFF to generate discrete templates~\cite{Raffel2020Exploring}. Other works~\cite{Shin2020AutoPrompt, han2021ptr, Scao2021How, Utama2021Avoiding} mine prompts from the training corpus based on heuristic rules/semantic relations. However, these methods are time-consuming for mining optimized prompts for target tasks. A series of methods are proposed to learn continuous/soft prompt embeddings, such as P-tuning~\cite{Xiao2021GPT}, P-tuning-V2~\cite{liu2021ptuningv2}, OptiPrompt~\cite{Zhong2021Factual}, Prefix-tuning~\cite{Li2021Prefix}. \citet{Zhao2021Discrete, Gu2021PPT} focus on the hybrid training with both discrete and continuous prompts. \citet{Hu2021Knowledgeable} consider the automatic expansion of label words and presents Knowledgeable Prompt-tuning (KPT) to utilize knowledge for the construction of verbalizers. 
\citet{Sun2021NSPBERT} and \citet{Wang2021Entailment} prompt the PLMs to make language inference in zero-shot learning. In addition, \citet{wang2021transprompt, Vu2021SPoT} consider transfer learning on continuous prompt-tuning. \citet{Li2021SentiPrompt, Chen20210KnowPrompt, Ma2021Template} focus on prompts for specific NLP tasks, such as sentiment analysis and information extraction.


Recently,~\citet{Wei2021Finetuned, Zhong2021Adapting, Min2021MetaICL, Mishra2021Reframing} tune PLMs on mixed data samples drawn from different NLP tasks with manually designed task-specific prompts. The resulting PLMs are then utilized to solve unseen tasks by zero-shot learning. 
These methods successfully work for large PLMs such as GPT-3~\cite{Brown2020Language} and T5~\cite{Raffel2020Exploring}, but consume a large amount of computation resources.
We further leverage data from non-target NLP tasks to make prompt-tuned PLMs have better capacities of adapting to unseen NLP tasks.

\section{Conclusion and Future Work}

In this paper, we present the~\emph{\fullmodel} framework (\emph{\model}) that enables better few-shot text classification for BERT-style models by explicitly capturing prompting semantics from non-target datasets.
Experiments show that~\emph{\model} consistently outperforms state-of-the-arts for prompt-based fine-tuning.
As for future work, we seek to extend~\emph{\model} to other tasks such as named entity recognition, text generation, and machine translation. In addition, we will explore continuous prompt-tuning for \emph{\model}.


\appendix
\balance 

\begin{table*}
\centering
\begin{small}
\begin{tabular}{l c c c c c c}
\midrule
\bf Group & \bf Category & \bf Task & \bf \#Training & \bf \#Testing & \bf $N$ & \bf Class Labels \\
\midrule
\multirow{3}{*}{Group 1: Sentiment} & \multirow{3}{*}{Single Sentence} & SST-2 & 6,920 & 872 & 2 & positive, negative \\
 & & MR & 8,662 & 2,000 & 2 & positive, negative \\
 & & CR & 1,775 & 2,000 & 2 & positive, negative \\
\midrule
\multirow{4}{*}{Group 2: NLI} & \multirow{4}{*}{Sentence Pair} & MNLI & 392,702 & 9,815 & 3 & entailment, neutral, contradiction \\
 & & SNLI & 549,367 & 9,842 & 3 & entailment, neutral, contradiction \\
 & & QNLI & 104,743 & 5,463 & 2 & entailment, not entailment \\
 & & RTE & 2,490 & 277 & 2 & entailment, not entailment \\
\midrule
\multirow{2}{*}{Group 3: Paraphrase} & \multirow{2}{*}{Sentence Pair} & MRPC & 3,668 & 408 & 2 & equivalent, not equivalent \\
 & & QQP & 363,846 & 40,431 & 2 & equivalent, not equivalent \\
\midrule
\end{tabular}
\end{small}
\caption{\label{tab:dataset}
Dataset statistics. We only sample $N\times K$ instances from the original training sets to form the few-shot training and development sets. The testing sets used in the experiments are full datasets.}
\end{table*}

\begin{table*}
\centering
\begin{tiny}
\begin{tabular}{ l | l | l | l}
\hline
\textbf{Task} & \textbf{Prompt} & \textbf{Option} & \textbf{Label word}  \\
\hline
\hline
\textbf{SST-2} & 
\makecell[l]{
\textbf{Template 1}: \texttt{[<s1>]}. It was \texttt{[MASK]}. \\
\textbf{Template 2}: \texttt{[<s1>]}. I thought it was \texttt{[MASK]}. \\
\textbf{Template 3}: \texttt{[<s1>]}. It is \texttt{[MASK]}. \\
\textbf{Template 4}: \texttt{[<s1>]}. The review is \texttt{[MASK]}. \\
\textbf{Template 5}: \texttt{[<s1>]}. A \texttt{[MASK]} one. \\
} &
\makecell[l]{
\textbf{Option 1}: Is \texttt{<x1>} or \texttt{<x2>}? \\
\textbf{Option 2}: Does \texttt{<x1>} or \texttt{<x2>}? \\
\textbf{Option 3}: \texttt{<x1>} or \texttt{<x2>}? \\
} &
\makecell[l]{
\textbf{Verbalizer 1}: Negative (Bad), Positive (Wonderful) \\
\textbf{Verbalizer 2}: Negative (Silly), Positive (Solid) \\
\textbf{Verbalizer 3}: Negative (Pathetic), Positive (Irresistible) \\
} \\
\hline
\textbf{MR} & 
\makecell[l]{
\textbf{Template 1}: \texttt{[<s1>]}. It was \texttt{[MASK]}. \\
\textbf{Template 2}: \texttt{[<s1>]}. A \texttt{[MASK]} piece of work. \\
\textbf{Template 3}: \texttt{[<s1>]}. It is \texttt{[MASK]}. \\
\textbf{Template 4}: \texttt{[<s1>]}. The film is \texttt{[MASK]}. \\
\textbf{Template 5}: \texttt{[<s1>]}. A really \texttt{[MASK]} movie. \\
} &
\makecell[l]{
\textbf{Option 1}: Is \texttt{<x1>} or \texttt{<x2>}? \\
\textbf{Option 2}: Does \texttt{<x1>} or \texttt{<x2>}? \\
\textbf{Option 3}: \texttt{<x1>} or \texttt{<x2>}? \\
} &
\makecell[l]{
\textbf{Verbalizer 1}: Negative (Horrible), Positive (Exquisite) \\
\textbf{Verbalizer 2}: Negative (Silly), Positive (Solid) \\
\textbf{Verbalizer 3}: Negative (Bad), Positive (Wonderful) \\
} \\
\hline
\textbf{CR} & 
\makecell[l]{
\textbf{Template 1}: \texttt{[<s1>]}. It was \texttt{[MASK]}. \\
\textbf{Template 2}: \texttt{[<s1>]}. It looks \texttt{[MASK]}. \\
\textbf{Template 3}: \texttt{[<s1>]}. It is \texttt{[MASK]}. \\
\textbf{Template 4}: \texttt{[<s1>]}. The quality is \texttt{[MASK]}. \\
\textbf{Template 5}: \texttt{[<s1>]}. I thought it was \texttt{[MASK]}. \\
} &
\makecell[l]{
\textbf{Option 1}: Is \texttt{<x1>} or \texttt{<x2>}? \\
\textbf{Option 2}: Does \texttt{<x1>} or \texttt{<x2>}? \\
\textbf{Option 3}: \texttt{<x1>} or \texttt{<x2>}? \\
} &
\makecell[l]{
\textbf{Verbalizer 1}: Negative (Horrible), Positive (Fantastic) \\
\textbf{Verbalizer 2}: Negative (Silly), Positive (Solid) \\
\textbf{Verbalizer 3}: Negative (Bad), Positive (Wonderful) \\
\textbf{Verbalizer 4}: Negative (Pointless), Positive (Neat) \\
} \\
\hline
\hline
\textbf{MNLI} & 
\makecell[l]{
\textbf{Template 1}: \texttt{[<s1>]}. You are right, \texttt{[MASK]}. \texttt{[<s2>]}.  \\
\textbf{Template 2}: \texttt{[<s1>]}. It was \texttt{[MASK]}. \texttt{[<s2>]}.  \\
\textbf{Template 3}: \texttt{[<s1>]}, \texttt{[<s2>]}. It is \texttt{[MASK]}. \\
\textbf{Template 4}: \texttt{[<s1>]}. It is true that \texttt{[MASK]}, \texttt{[<s2>]}.  \\
\textbf{Template 5}: \texttt{[<s1>]}. \texttt{[MASK]}. Then, \texttt{[<s2>]}.  \\
} &
\makecell[l]{
\textbf{Option 1}: Is \texttt{<x1>} or \texttt{<x2>} or\texttt{<x3>} ? \\
\textbf{Option 2}: Based on the paragraph above, is the \\ following \texttt{<x1>} or \texttt{<x2>} or \texttt{<x3>}? \\
} &
\makecell[l]{
\textbf{Verbalizer 1}: Contradiction (Next), \\ Entailment (Exactly), Neutral (Indeed) \\
\textbf{Verbalizer 2}: Contradiction (Wrong), \\ Entailment (True), Neutral (Uncertain) \\
\textbf{Verbalizer 3}: Contradiction (Otherwise), \\ Entailment (Fine), Neutral (Plus) \\
\textbf{Verbalizer 4}: Contradiction (Otherwise), \\ Entailment (Exactly), Neutral (Naturally) \\
} \\
\hline
\textbf{SNLI} & 
\makecell[l]{
\textbf{Template 1}: \texttt{[<s1>]}. \texttt{[MASK]}, no, \texttt{[<s2>]}.   \\
\textbf{Template 2}: \texttt{[<s1>]}. \texttt{[MASK]}, in this case, \texttt{[<s2>]}.  \\
\textbf{Template 3}: \texttt{[<s1>]}. \texttt{[MASK]}, I think, \texttt{[<s2>]}.  \\
\textbf{Template 4}: \texttt{[<s1>]}, \texttt{[<s2>]}. It was \texttt{[MASK]}. \\
\textbf{Template 5}: \texttt{[<s1>]}. \texttt{[MASK]}, \texttt{[<s2>]}. \\
} &
\makecell[l]{
\textbf{Option 1}: Is \texttt{<x1>} or \texttt{<x2>} or\texttt{<x3>} ? \\
\textbf{Option 2}: Based on the paragraph above, is the \\ following \texttt{<x1>} or \texttt{<x2>} or \texttt{<x3>}? \\
} &
\makecell[l]{
\textbf{Verbalizer 1}: Contradiction (Next), \\ Entailment (Exactly), Neutral (Indeed) \\
\textbf{Verbalizer 2}: Contradiction (Wrong), \\ Entailment (True), Neutral (Uncertain) \\
\textbf{Verbalizer 3}: Contradiction (Instead), \\ Entailment (Indeed), Neutral (Basically) \\
\textbf{Verbalizer 4}: Contradiction (Except), \\ Entailment (Alright), Neutral (Watch) \\
} \\
\hline
\textbf{QNLI} & 
\makecell[l]{
\textbf{Template 1}: Question: \texttt{[<s1>]}? \texttt{[<s2>]}. The answer: \\ \texttt{[MASK]}.  \\
\textbf{Template 2}: Question: \texttt{[<s1>]}? \texttt{[<s2>]}. \texttt{[MASK]}.  \\
\textbf{Template 3}: Question: \texttt{[<s1>]}? \texttt{[MASK]}, Yes, \texttt{[<s2>]}.  \\
\textbf{Template 4}: \texttt{[<s1>]}? \texttt{[MASK]}, it is known that \texttt{[<s2>]}.  \\
\textbf{Template 5}: \texttt{[<s1>]}? \texttt{[MASK]}. Then, \texttt{[<s2>]}.  \\
} &
\makecell[l]{
\textbf{Option 1}: Is \texttt{<x1>} or \texttt{<x2>} ? \\
\textbf{Option 2}: Based on the question, is the \\ following \texttt{<x1>} or \texttt{<x2>}? \\
\textbf{Option 3}: Is the answer \texttt{<x1>} or \texttt{<x2>}? \\
} &
\makecell[l]{
\textbf{Verbalizer 1}: Entailment (Yes), Not Entailment (No) \\
\textbf{Verbalizer 2}: Entailment (Okay), \\ Not Entailment (Nonetheless) \\
\textbf{Verbalizer 3}: Entailment (Notably), Not Entailment (Yet) \\
} \\
\hline
\textbf{RTE} & 
\makecell[l]{
\textbf{Template 1}: \texttt{[<s1>]}. \texttt{[<s2>]}. The answer: \texttt{[MASK]}.  \\
\textbf{Template 2}: \texttt{[<s1>]}. \texttt{[<s2>]}. \texttt{[MASK]}.  \\
\textbf{Template 3}: \texttt{[<s1>]}. \texttt{[MASK]}, I think, \texttt{[<s2>]}.  \\
\textbf{Template 4}: \texttt{[<s1>]}. The question: \texttt{[<s2>]}? It is \texttt{[MASK]}.  \\
\textbf{Template 5}: \texttt{[<s1>]}. \texttt{[MASK]}. I believe, \texttt{[<s2>]}.  \\
} &
\makecell[l]{
\textbf{Option 1}: Is \texttt{<x1>} or \texttt{<x2>} ? \\
\textbf{Option 2}: Based on the question, the answer \\ is \texttt{<x1>} or \texttt{<x2>}? \\
\textbf{Option 3}: Is the answer \texttt{<x1>} or \texttt{<x2>}? \\
} &
\makecell[l]{
\textbf{Verbalizer 1}: Entailment (So), \\ Not Entailment (Meanwhile) \\
\textbf{Verbalizer 2}: Entailment (Yes), Not Entailment (No) \\
\textbf{Verbalizer 3}: Entailment (Notably), Not Entailment (Yet) \\
} \\
\hline
\hline
\textbf{MRPC} & 
\makecell[l]{
\textbf{Template 1}: \texttt{[<s1>]}. \texttt{[<s2>]}. The answer: \texttt{[MASK]}.  \\
\textbf{Template 2}: \texttt{[<s1>]}. \texttt{[<s2>]}. \texttt{[MASK]}.  \\
\textbf{Template 3}: \texttt{[<s1>]}. \texttt{[MASK]}, however, \texttt{[<s2>]}.  \\
\textbf{Template 4}: \texttt{[<s1>]}. \texttt{[<s2>]}. In fact \texttt{[MASK]}.  \\
\textbf{Template 5}: \texttt{[<s1>]}. \texttt{[MASK]}. that's right, \texttt{[<s2>]}.  \\
} &
\makecell[l]{
\textbf{Option 1}: Is \texttt{<x1>} or \texttt{<x2>} ? \\
\textbf{Option 2}: Are two question \texttt{<x1>} or \texttt{<x2>}? \\
\textbf{Option 3}: \texttt{<x1>} or \texttt{<x2>}? \\
} &
\makecell[l]{
\textbf{Verbalizer 1}: 0 (Alas), 1 (Rather) \\
\textbf{Verbalizer 2}: 0 (Different), 1 (Same) \\
\textbf{Verbalizer 3}: 0 (Wrong), 1 (Right) \\
} \\
\hline
\textbf{QQP} & 
\makecell[l]{
\textbf{Template 1}: \texttt{[<s1>]}. \texttt{[<s2>]}. The answer: \texttt{[MASK]}.  \\
\textbf{Template 2}: \texttt{[<s1>]}. \texttt{[<s2>]}. \texttt{[MASK]}.  \\
\textbf{Template 3}: \texttt{[<s1>]}. \texttt{[MASK]}, however, \texttt{[<s2>]}.  \\
\textbf{Template 4}: \texttt{[<s1>]}. \texttt{[<s2>]}. In fact \texttt{[MASK]}.  \\
\textbf{Template 5}: \texttt{[<s1>]}. \texttt{[MASK]}. that's right, \texttt{[<s2>]}.  \\
} &
\makecell[l]{
\textbf{Option 1}: Is \texttt{<x1>} or \texttt{<x2>} ? \\
\textbf{Option 2}: Are two question \texttt{<x1>} or \texttt{<x2>}? \\
\textbf{Option 3}: \texttt{<x1>} or \texttt{<x2>}? \\
} &
\makecell[l]{
\textbf{Verbalizer 1}: 0 (Alas), 1 (Rather) \\
\textbf{Verbalizer 2}: 0 (Different), 1 (Same) \\
\textbf{Verbalizer 3}: 0 (Wrong), 1 (Right) \\
} \\
\hline
\end{tabular}
\end{tiny}
\caption{\label{tab:pov_list} The Prompts, Options and Verbalizers (POV) for each task. \texttt{<s1>} and \texttt{<s2>} denote the input sentences. \texttt{<x1>}, \texttt{<x2>} and \texttt{<x3>} denote the label words.}
\end{table*}

\section{Dataset Statistics}

In the main experiments, we employ 9 different NLP datasets for evaluation. As shown in Table~\ref{tab:dataset}, we divided all datasets into three groups, i.e., Sentiment, NLI, and Paraphrase. During multi-task training, we select two groups of tasks with full training data for~\emph{POV} prompt-tuning with the auxiliary \emph{KSMLM} objective. After that, we prompt-tune the model over the target task in the few-shot learning setting. The corresponding group of the target task is unseen during multi-task training.

\begin{table*}[ht]
\centering
\begin{small}
\resizebox{\textwidth}{!}{
\begin{tabular}[\textwidth]{c | c | ccccccc | c}
\midrule
\bf Paradigm &\bf  Method &\bf  AX-b &\bf  AX-g &\bf  BoolQ &\bf  CB &\bf  SST-5 &\bf  TREC &\bf  Subj &\bf  Avg. \\
\midrule
FT & Fine-tuning & 47.51\scriptsize $\pm$1.8 & 60.83\scriptsize $\pm$2.0 & 65.96\scriptsize $\pm$1.3 & 73.21\scriptsize $\pm$1.3 & 40.10\scriptsize $\pm$3.4 & 59.30\scriptsize $\pm$1.8 & 73.00\scriptsize $\pm$2.0 & 59.99 \\
\midrule
\multirow{4}{*}{PT} & PET & 60.28\scriptsize $\pm$1.2 & 64.08\scriptsize $\pm$0.8 & 70.54\scriptsize $\pm$1.6 & 82.09\scriptsize $\pm$2.0 & 44.10\scriptsize $\pm$1.7 & 84.90\scriptsize $\pm$1.9 & 89.30\scriptsize $\pm$1.0 & 70.76 \\
& LM-BFF & 61.53\scriptsize $\pm$1.4 & 63.89\scriptsize $\pm$1.9 & 71.30\scriptsize $\pm$2.1 & 82.14\scriptsize $\pm$2.6 & 46.10\scriptsize $\pm$1.3 & 84.80\scriptsize $\pm$1.5 & 89.25\scriptsize $\pm$1.0 & 71.29 \\
& P-Tuning & 62.23\scriptsize $\pm$0.8 & 63.19\scriptsize $\pm$1.2 & 72.88\scriptsize $\pm$0.9 & 83.08\scriptsize $\pm$1.8 & 48.20\scriptsize $\pm$1.5 & 85.10\scriptsize $\pm$1.9 & 89.35\scriptsize $\pm$1.1 & 72.00 \\
& \emph{\model} & \bf 64.25\scriptsize $\pm$1.2 & \bf 69.44\scriptsize $\pm$1.4 & \bf 74.06\scriptsize $\pm$1.6 & \bf 83.92\scriptsize $\pm$0.9 & \bf 48.35\scriptsize $\pm$1.0 & \bf 85.90\scriptsize $\pm$0.8 & \bf 90.15\scriptsize $\pm$1.2 & \bf 73.72 \\
\midrule
\end{tabular}
}
\end{small}
\caption{\label{tab:general_results2} Additional experiments for comparison between~\emph{\model} and baselines over all testing sets in terms of accuracy (\%) and standard deviation.
}
\end{table*}

\begin{table}
\centering
\begin{small}
\resizebox{\linewidth}{!}{
\begin{tabular}{l | ccc}
\midrule
\textbf{Method}/\textbf{Group} & \textbf{Group 1} & \textbf{Group 2} & \textbf{Group 3}\\
\midrule
 \emph{\model} & \textbf{91.0} & \textbf{70.2} & \textbf{77.9}\\
 \emph{\model} w/o. \emph{KSMLM} & 90.9 & 69.1 & 73.7 \\
\midrule
 MLM & 87.1 & 67.4 & 72.0 \\
 \emph{KSMLM} (w/o. \emph{OKR}) & 90.7 & 69.9 & 76.8 \\
 \emph{KSMLM} (w/o. Options) & 90.1 & 68.2 & 76.3 \\
 \emph{KSMLM} (w/o. Verbalizer) & 85.0 & 62.4 & 66.7 \\
\midrule
\end{tabular}
}
\end{small}
\caption{\label{tab:KSMLM_analysis} The ablation analysis of the \emph{KSMLM} task in terms of accuracy (\%).
}
\end{table}

\begin{table}
\centering
\begin{small}
\begin{tabular}{l | ccc | c}
\midrule
\textbf{Paradigm}/\textbf{Task} & \textbf{SST-2} & \textbf{MR} & \textbf{CR} & \textbf{Avg.}\\
\midrule
 \emph{POV} & \textbf{92.9} & 87.7 & \textbf{91.8} & \textbf{90.8}\\
\midrule
 Multiple-choice & 82.7 & 73.9 & 80.9 & 79.2\\
 Yes/No & 92.6 & \textbf{87.9} & 91.6 & 90.7\\
\midrule
\end{tabular}
\end{small}
\caption{\label{tab:paradigm_comparison} The comparison between different paradigms in terms of accuracy (\%).
}
\end{table}

\section{\emph{POV} Examples}

As shown in Table~\ref{tab:pov_list}, we list the designed \emph{POV}s for all the tasks. Note that for each task group, the options are the same, but verbalizers of these tasks may be different. For example, SST-2, MR, and CR have the same schema of options, but with different verbalizers. 

\section{Detailed Experiments for the \emph{KSMLM} Task}

We further conduct experiments over each group to evaluate the effectiveness of different settings in \emph{KSMLM}. The baselines for comparison include:
\begin{itemize}
    \item \textbf{\emph{\model} w/o. \emph{KSMLM}}: It means the training process on source tasks without the \emph{KSMLM} learning objective before prompt-tuning over the target task.
    
    \item \textbf{MLM}: It means that we directly train the vanilla MLM based on the full training data from source tasks.
    
    \item \textbf{\emph{KSMLM} (w/o. \emph{OKR})}: It means that we randomly select options without the K-Means algorithm and the knowledge-guided options construction process.
    
    \item \textbf{\emph{KSMLM} (w/o. Options)}: It means that we directly remove the options in \emph{POV}.
    
    \item \textbf{\emph{KSMLM} (w/o. Verbalizer)}: It means that the prediction search space at each masked position is the whole BERT vocabulary rather than the designed limited collection of label words (expressed by options).
\end{itemize}

As shown in Table~\ref{tab:KSMLM_analysis}, we follow the same settings with the ablation study in Table \ref{tab:ablation} to report the mean accuracy values of each group. We can draw the following conclusions: 1) Compared to vanilla MLM, the results indicate that \emph{KSMLM} is an irreplaceable task for the improvement of the model generalization power. 2) We also find that if we ignore the verbalizer construction, the results decrease to a large degree, and lower than \emph{\model} w/o. \emph{KSMLM}. It means that verbalizers are crucial for template-based prompt-tuning. 3) When \emph{OKR} or options are removed, the results also decline, indicating the effectiveness of these techniques.

\section{Comparing \emph{POV} with Other Paradigms}

To compare the proposed \emph{POV} paradigm with other paradigms, we perform experiments over SST-2, MR, and CR tasks. The alternative paradigms are as follows:

\begin{itemize}
    \item \textbf{Multiple-choice}. It is a unified template to list all the candidate results. For example, an input sentence can be ``The Disney cartoons are very interesting for children to enrich their extracurricular life. A. great; B. terrible. It is \texttt{[MASK]}.''. This paradigm is closely in line with PPT~\cite{Gu2021PPT}.
    
    \item \textbf{Yes/No}. We can reformulate the multi-class classification tasks into a series of binary classification. Take NLI for example. We can design three templates for each class, i.e ``Are these descriptions are entailment?'', ``Are these descriptions are neutral?'', and ``Are these descriptions are contradiction?''. We follow~\citet{Zhong2021Adapting} to add an MLP layer on the top of the PLM to obtain the output of the \texttt{[MASK]} token to classify the answer to be ``Yes'' or ``No''.
\end{itemize}

Experimental results in Table~\ref{tab:paradigm_comparison} show that in average, \emph{POV} outperforms all baselines. For Multi-choice, we find the results decline a lot. We guess that the PLM is hard to understand and generate the items number, such as ``A, B, C, D''. In addition, we find the paradigm ``Yes/No'' has a similar performance to \emph{POV}. Overall, the experiments prove the effectiveness of \emph{POV}, which is easy to implement and avoids the transformation to multiple binary classification tasks for tasks with multiple classes. 



\section{Additional Evaluation Results over Other Tasks}
In this part, we further present additional experiments over other tasks from GLUE~\cite{Wang2019GLUE} and SuperGLUE~\cite{Wang2019SuperGLUE}, including AX-b, AX-g, BoolQ, CB, SST-5, TREC and Subj. The data statistics can be found in the original papers.  We choose standard fine-tuning, PET~\cite{Schick2021Exploiting}, LM-BFF~\cite{Xiao2021GPT} as our baselines to make comparison. In this experiment, we only conduct task-specific single-task learning to evaluate the efficiency of the \emph{POV} paradigm. We also set $K=16$. As shown in Table~\ref{tab:general_results2}, we can draw the following conclusions. 1) Our \emph{\model} framework outperforms strong baselines over these tasks. 2) SST-5 and TREC are challenging tasks with many labels, which consist of 5 and 6 classes, respectively. Experiments show that our proposed \emph{POV} paradigm can also achieve the best performances over this scenario.

\end{document}